\documentclass{llncs}
\usepackage{makeidx}  

\usepackage{verbatim}
\usepackage{amsfonts}
\usepackage{nccmath}
\usepackage[algo2e,ruled,vlined]{algorithm2e}
\usepackage{caption}
\usepackage{subfigure}
\usepackage{stackengine}
\usepackage{multirow}

\usepackage{epstopdf}

\usepackage{floatrow}

\usepackage{xcolor}
\usepackage[export]{adjustbox}

\newcommand{\figViz}[1]{\includegraphics[height=2.8cm]{#1}}

\title{\Large Multi-modal analysis of genetically-related subjects using SIFT descriptors in brain MRI}

\titlerunning{*** Multi-modal SIFT analysis}  
\author{Kuldeep Kumar$^{1,2}$, Laurent Chauvin$^{1}$, Mathew Toews$^{1}$, \\ Olivier Colliot$^{2,3}$ \and Christian Desrosiers$^{1}$}

\institute{$^{1}$LIVIA, \'Ecole de technologie sup\'erieure (\'ETS), Montreal, Canada \\
	$^{2}$Aramis project-team, Inria Paris, Sorbonne Universit\'es, UPMC Univ Paris 06, Inserm, CNRS, Institut du cerveau et la moelle (ICM) - H\^opital Piti\'e-Salp\^etri\`ere, Boulevard de l$\rq$h\^opital, F-75013, Paris, France \\
	$^{3}$AP-HP, Departments of Neurology and Neuroradiology, H\^opital Piti\'e-Salp\^etri\`ere, 75013, Paris, France}

\begin{document}
	
\maketitle
\begin{abstract}
So far, fingerprinting studies have focused on identifying features from single-modality MRI data, which capture individual characteristics in terms of brain structure, function, or white matter microstructure. However, due to the lack of a framework for comparing across multiple modalities, studies based on multi-modal data remain elusive. This paper presents a multi-modal analysis of genetically-related subjects to compare and contrast the information provided by various MRI modalities. The proposed framework represents MRI scans as bags of SIFT features, and uses these features in a nearest-neighbor graph to measure subject similarity. Experiments using the T1/T2-weighted MRI and diffusion MRI data of 861 Human Connectome Project subjects demonstrate strong links between the proposed similarity measure and genetic proximity. 
\end{abstract}

\section{Introduction}

The human brain shows a significant amount of variability in terms of structure \cite{mangin2004framework}, function \cite{mueller2013individual}, and white matter architecture \cite{de2011atlasing}. Recently, various studies have focused on characterizing this variability using brain \emph{fingerprints}, for instance, based on shape \cite{wachinger2015brainprint}, functional connectivity \cite{finn2015functional}, white matter fiber geometry \cite{KUMAR2017242} or voxel-wise diffusion density \cite{yeh2016connectometry}. The importance of these fingerprint studies is motivated by the fact that brain characteristics are largely determined by genetic factors \cite{thompson2013genetics}, and that various neurological disorders like Parkinson \cite{geevarghese2015subcortical} and autism \cite{goldman2013motor} have been linked to brain abnormalities.    

Most fingerprint studies use magnetic resonance imaging (MRI) data, which provides a non-invasive way to probe the structure, function and white matter connectivity of the brain. Studies focusing on structural MRI modalities like T1- or T2-weighted images derive their fingerprints mainly based on morphometry. In \cite{wachinger2015brainprint}, Wachinger et al. quantify the shape of cortical and sub-cortical structures via the spectrum of the Laplace-Beltrami operator. The resulting representation, called Brainprint, is used for subject identification and analyzing potential genetic influences on brain morphology. In \cite{toews2010feature}, Toews et al. represent images as a collection of localized image descriptors, and apply scale-space theory to analyze their distribution at the characteristic scale of underlying anatomical structures. This morphometry-based representation is employed to identify distinctive anatomical patterns of genetically-related individuals or subjects with a known brain disease.

Various methods have also been proposed to characterize brain properties derived from diffusion MRI (dMRI) or functional MRI (fMRI). In \cite{kochunov2015heritability}, Kochunov et al. analyze heritability of fractional anisotropy (FA) in white matter and compare findings across subjects from different datasets. Likewise, Yeh et al. \cite{yeh2016connectometry} build a local connectome fingerprint using dMRI data, based on the diffusion density at each voxel, and use their proposed fingerprint to analyze genetically-related subjects. In \cite{finn2015functional}, Finn et al. consider the correlation between time courses of atlas-defined nodes to generate a functional connectivity profile, and use this profile to identify individuals across scan sessions, both for task and rest conditions. Moreover, Miranda et al. \cite{miranda2014connectotyping} propose a linear model to describe the activity of brain regions in resting-state fMRI as a weighted sum of its functional neighboring regions. Their functional fingerprint, derived from the model's coefficients, has the ability to predict individuals using a limited number of non-sequential frames. 


So far, brain fingerprinting studies have used single-modality imaging data to characterize the structural, functional or white-matter connectivity profiles of individuals. Combining multi-modal data in a single fingerprint would provide a more discriminative characterization of individuals and help understanding the unique contribution of each modality in this characterization. However, such multi-modal fingerprinting analysis has been elusive, due to the lack of a common framework for comparing subjects across modalities. Motivated by this, we propose a multi-modal analysis approach that combines information from structural MRI (T1 and T2) and diffusion MRI into a single framework. The idea is to represent each image using local features and approximate image manifold by computing pairwise subject similarity in an efficient way. Building from the work presented in \cite{7493398}, this framework represents images as a bags of features (BoF), where features are defined based on \emph{scale-invariant feature transform} (SIFT) descriptors. This technique has the advantage of being alignment independent and, thus, facilitates comparisons across different subjects. To find similar subjects, we first build a nearest-neighbor (NN) graph by matching SIFT descriptors, and then compute the pairwise geodesic distance between BoFs using the Jaccard distance metric. In this work, we extend the framework described in \cite{7493398} to include T1-weighted (without skull), T2-weighted, and diffusion-weighted MRI, and present a comprehensive quantitative and qualitative analysis using $861$ subjects from Human Connectome Project. To our knowledge, this study is the first step towards using multi-modal information for subject fingerprinting and analyzing genetically-related subjects. 


\section{Materials and Methods}
\label{sec:MaterialsAndMethods}

We start by describing the data and preprocessing steps of our study. In Section \ref{subsec:similarity}, we then present the proposed approach for comparing subjects using multi-modal data. 

\subsection{Data and preprocessing}
\label{subsec:DataAndPreProc}

We have used the pre-processed structure and diffusion MRI of 861 subjects (482 females, 378 male and 1 unknown, age 22-35) from the Q3 release of the Human Connectome Project \cite{van2013wu}. These images were acquired on a Siemens Skyra $3$T scanner and pre-processed 
as described in \cite{glasser2013minimal}. Further details can be obtained from HCP Q3 data release manual\footnote{\url{http://www.humanconnectome.org/documentation/Q3/}}. 

For structural MRI we considered high-resolution T1-weighted ($0.7$ mm), high-resolution T2-weighted ($0.7$ mm), and Freesurfer processed T1-weighted (skull stripped, $1$ mm). In the case of dMRI data, signal reconstruction was performed with the freely available DSI Studio toolbox \cite{yeh2010generalized}. All subjects were reconstructed in MNI space (at $1$ mm resolution) using the Q-Space diffeomorphic reconstruction (QSDR) option in DSI Studio. Three diffusivity measures were extracted to characterize white matter micro-structure: Generalized Fractional Anisotropy (GFA) and Normalized Quantitative Anisotropy (NQA0 and NQA1). GFA extends the standard FA measure to orientation distribution functions (ODF) based on spherical harmonics. NQA is a scaled version of quantitative anisotropy, which is calculated from the peak orientations on a spin distribution function (SDF). More information about these measures can be found in \cite{yeh2010generalized}.  

\subsection{Multi-modal subject similarity}
\label{subsec:similarity}

Computing similarities between pairs of subjects, based on their multi-modal data, involves multiple steps. In the first step, 3D keypoints are located in the scans of each subject by finding the local extrema (i.e., maxima or minima) of the difference of Gaussians (DoG) occurring at multiple scales. Keypoints with low contrast or corresponding to edge response are discarded, and remaining ones encoded into a feature vector (i.e, the descriptor) using the histogram of oriented gradients (HOG) within a small neighborhood. Note that these descriptors are robust to changes in illumination, scale and rotation, and are thus efficient for comparing images acquired using different scanners or imaging parameters. Each subject is then represented as an orderless bag of features (BoF), containing all the descriptors found in this subject's scans. This representation provides a simple and extensible way of incorporating data from multiple modalities.

Because the BoFs of two subjects may contain different numbers of descriptors, they cannot be directly compared. To circumvent this problem, we construct an intrinsic manifold of subject appearance using a feature-to-feature nearest-neighbor (NN) graph. In this graph, each descriptor is represented by a node and is connected to its $K$ most similar descriptors based on Euclidean distance. Let $A$ and $B$ be the BoF of two subjects, the similarity between these subjects can then be evaluated directly via the Jaccard measure:
\begin{equation}\label{eq:jaccard}
    J(A,B) \ = \ \frac{|A \cap B|}{|A| + |B| - |A \cap B|},
\end{equation}
where $|A \cap B|$ is the number of edges between nodes in $A$ and those in $B$. 

\section{Experiments and results}
\label{sec:Results}


We use the MotherID, Age, TwinStat, and Zygosity fields of the HCP twin data to obtain $84$ mono-zygotic (MZ) twin pairs, $84$ di-zygotic (DZ) twin pairs, and $217$ non-twin sibling subjects (NT). The proposed method is applied on each type of structure or diffusion image (i.e., T1, T2, GFA, NQA0, NQA1) to identify pairs of genetically-related subjects. For simplification purposes, we refer to those image types as \emph{modalities}. These modalities are compared using the recall@$k$ measure, which is the proportion of genetically-related subjects in the set of $k$ most similar subjects, as defined in Eq. (\ref{eq:jaccard}). For a given sibling type, recall@$k$ values are computed for all subjects having at least one sibling of this type, and averaged across these subjects. We then asses the complementarity of different modalities, by using them together for twin/sibling identification. A qualitative analysis on a single non-twin sibling pair is also performed to visualize the location and scale of features matches.

Parameter $K$ determines the number of connections in the nearest-neighbor graph, which can impact the proposed similarity measure. In our experiments, we tested $K \in \{10, 20, \ldots, 50\}$ and found nearly the same results for all tested values. In what follows, we report results obtained with $K=20$.

\subsection{Identification of genetically-related subjects}

\begin{figure}[ht]
	{\centering
		\mbox{
			\figViz{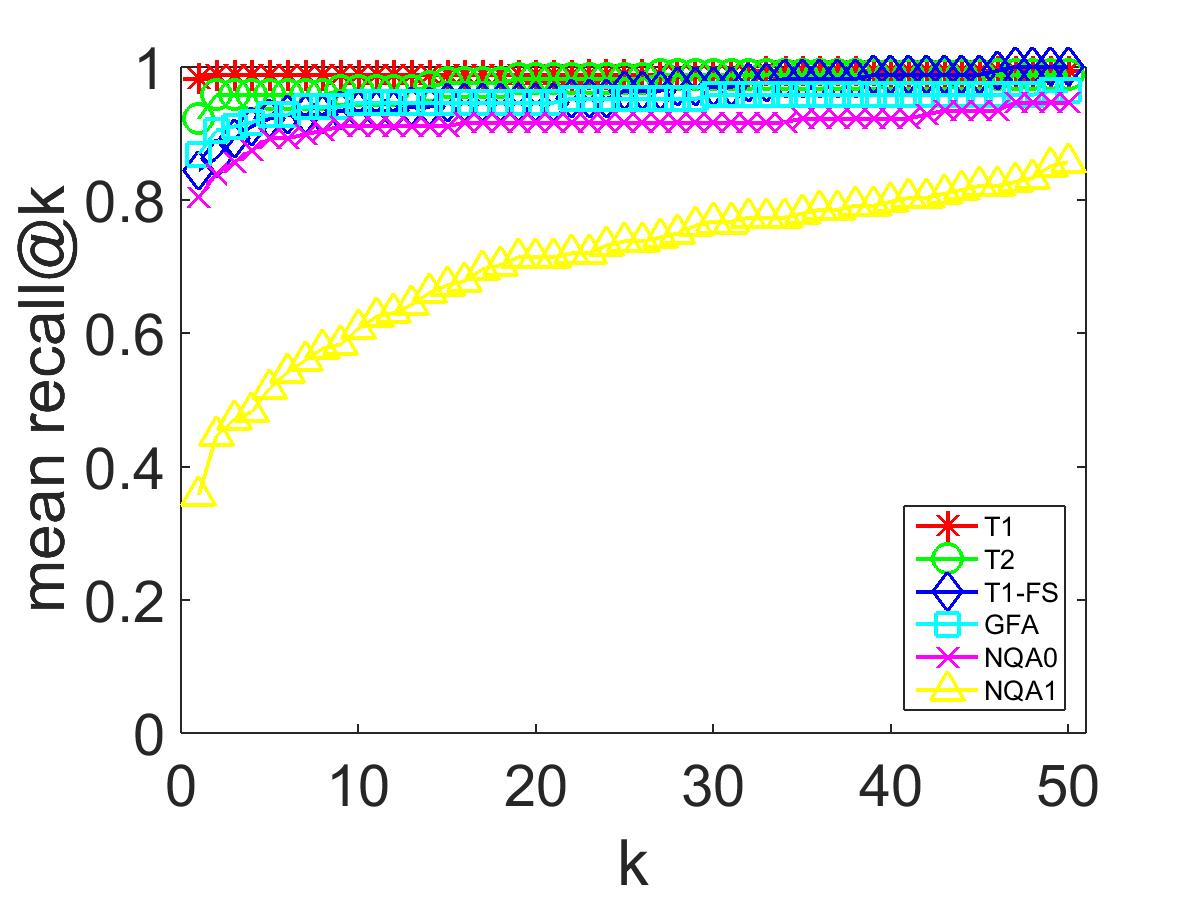}
			\hspace{0.01mm}
			\figViz{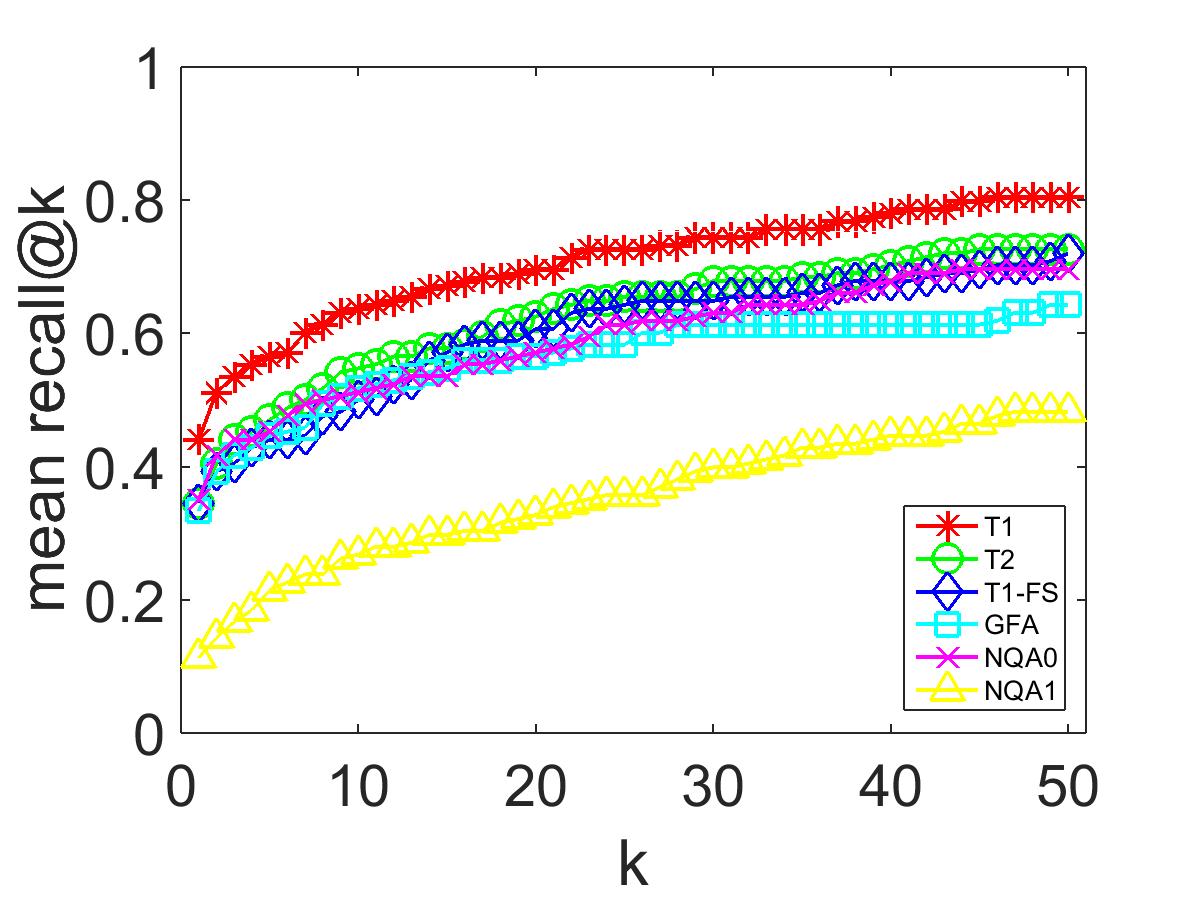}
			\hspace{0.01mm}
			\figViz{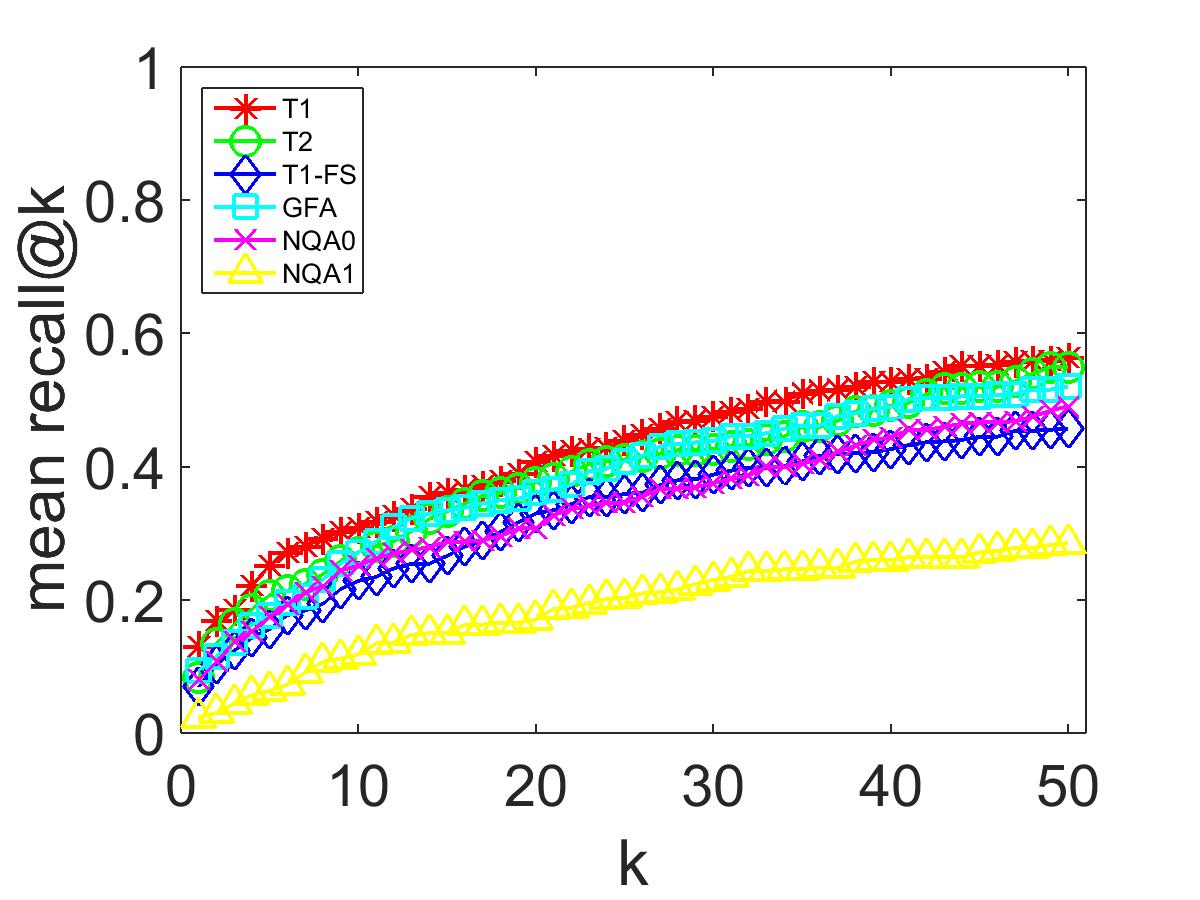}
		}
		\mbox{
			\shortstack{\figViz{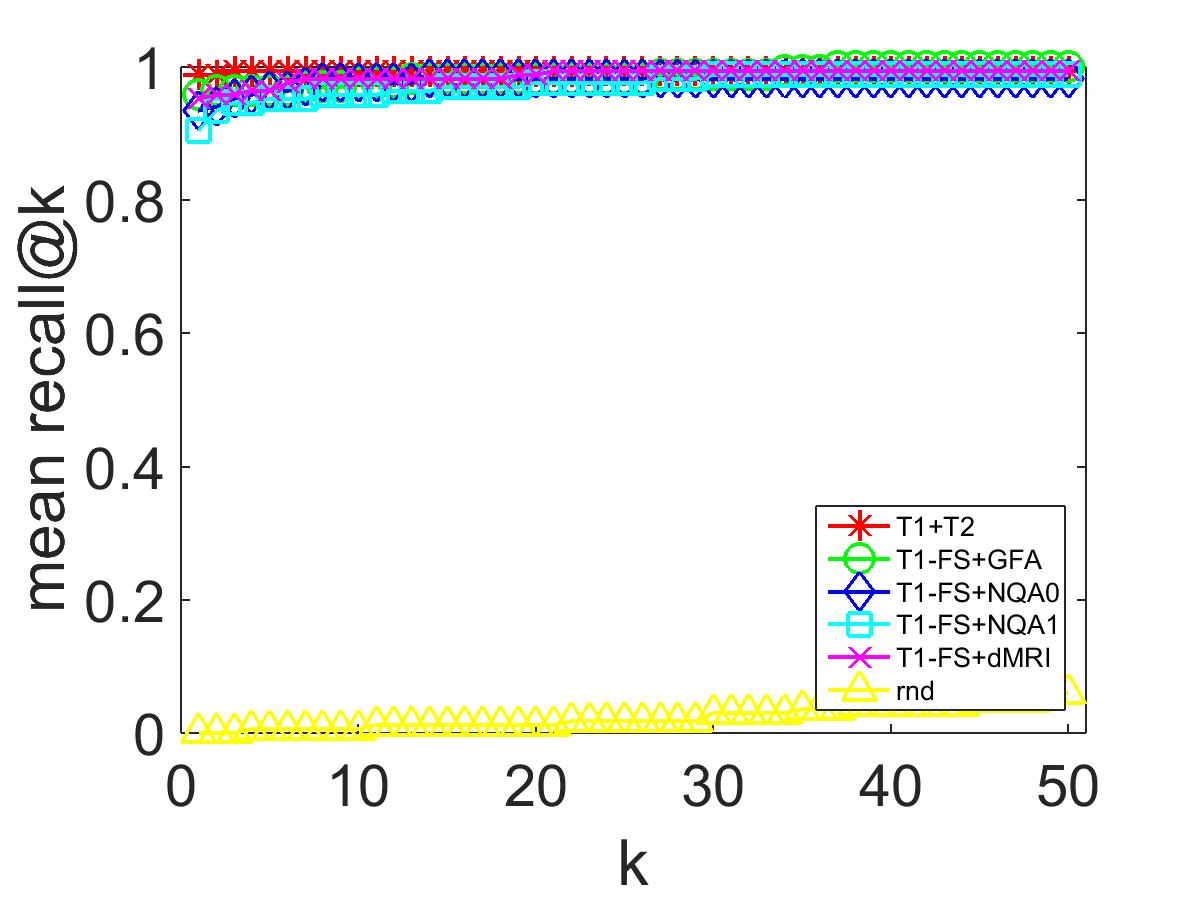} \\[1mm] {\small MZ}}
			\hspace{0.01mm}
			\shortstack{\figViz{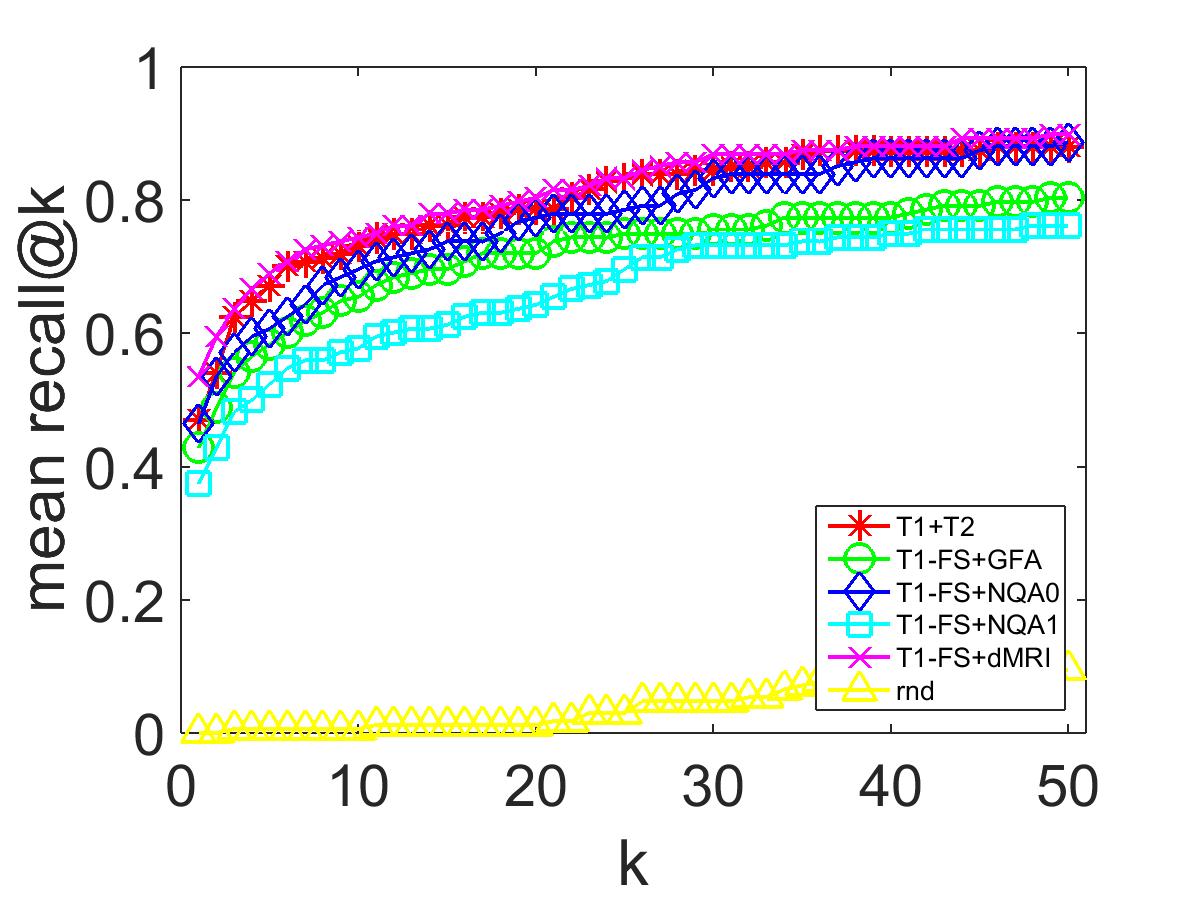} \\[1mm] {\small DZ}}
			\hspace{0.01mm}
			\shortstack{\figViz{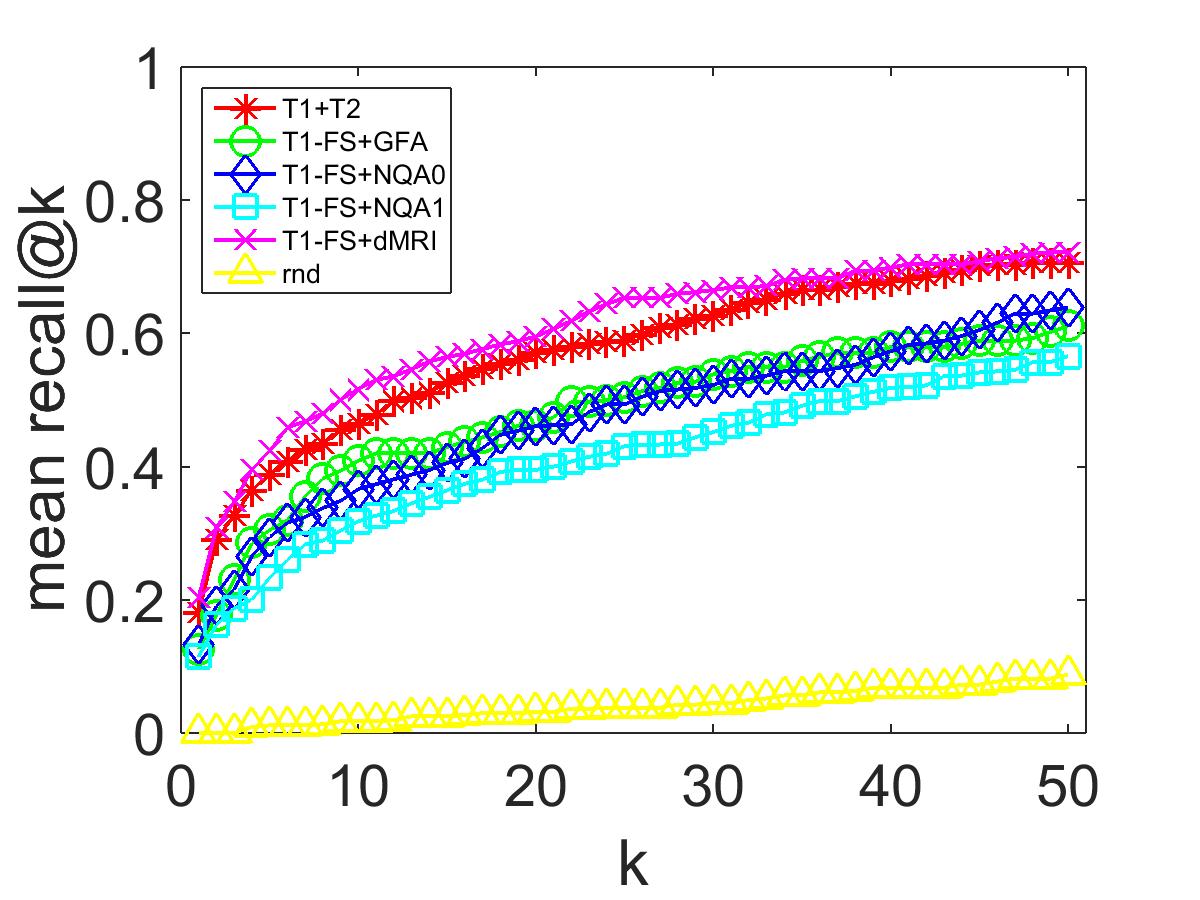} \\[1mm] {\small NT}}
		}
		
		\caption{Mean recall@$k$ for MZ, DZ and NT sibling identification using single-modality (top row) multi-modality data (bottom row).}
		\label{fig:Identify_Twin}
	}
\end{figure}

Figure \ref{fig:Identify_Twin} shows the mean recall@$k$, $k=1,\dots,50$, obtained for identifying MZ, DZ and NT siblings. Note that we perform the task of sibling identification over all $861$ subjects. The top row gives results obtained using single-modality data. T1-FS represents lower-resolution T1 images pre-processed with FreeSurfer. These images were included to evaluate the impact of scan resolution and for combining structural and diffusion data. The bottom row gives results obtained using a combination of two different modalities, except for T1+dMRI which corresponds to T1-FS combined with all diffusion measures (i.e., GFA, NQA0, NQA1). To consider the chance factor, we also provide the recall@$k$ resulting from a random subject similarity measure (denoted as \emph{rnd} in the bottom row plots).

\begin{figure}[h]
	{\centering
		
		\mbox{
			\shortstack{\includegraphics[height=2.8cm]{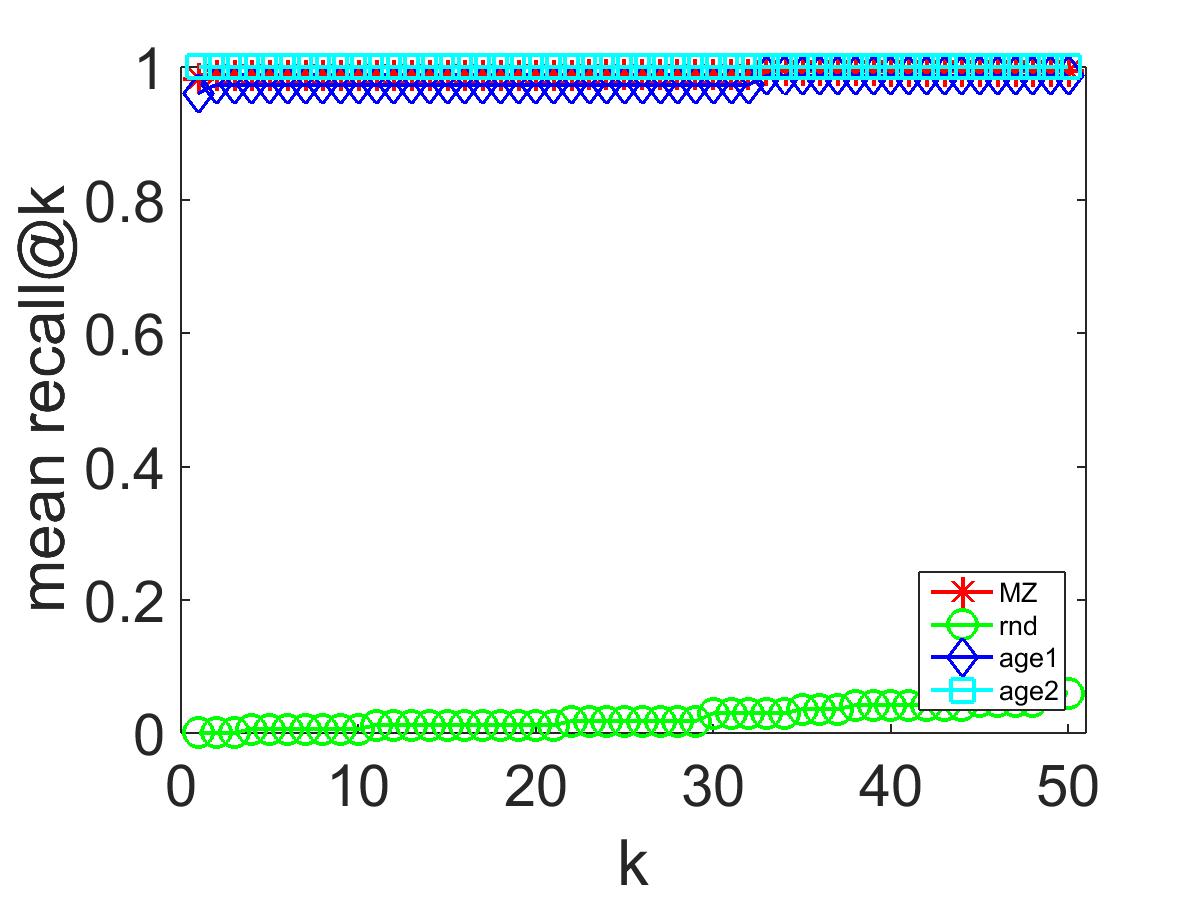}
				\\[1mm] {MZ}}
			\hspace{0.01mm}
			\shortstack{\includegraphics[height=2.8cm]{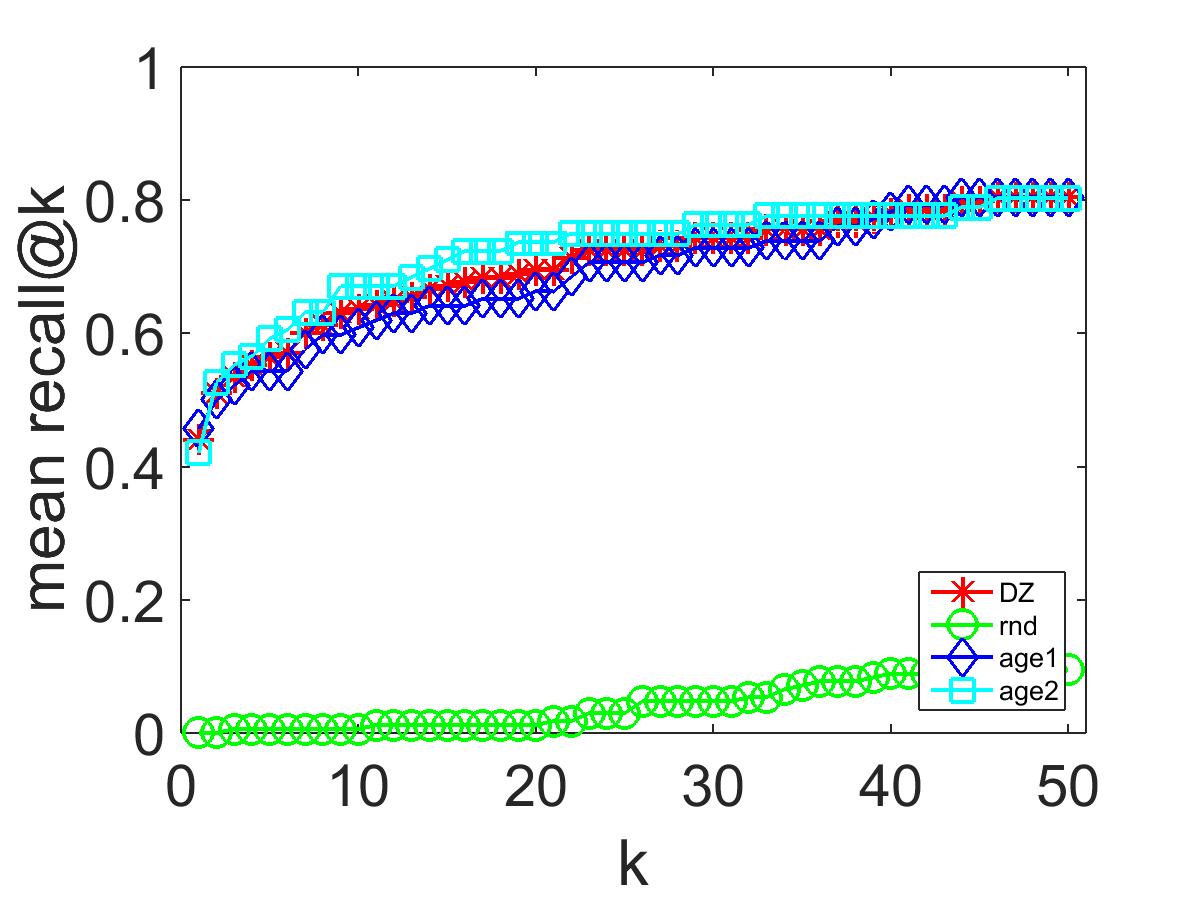}
				\\[1mm] {DZ}}
			\hspace{0.01mm}
			\shortstack{\includegraphics[height=2.8cm]{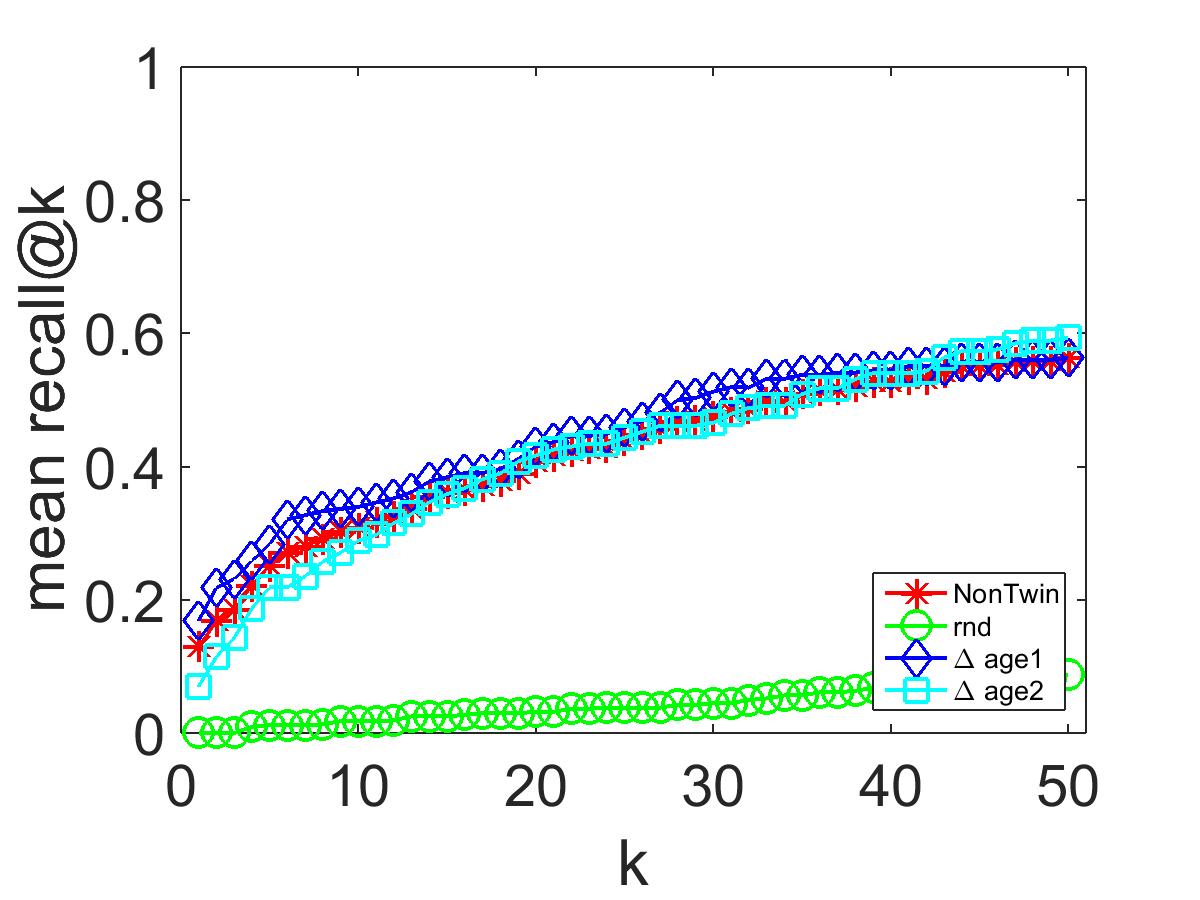}
				\\[1mm] {NT}}
		}
		
		\caption{Age impact on Recall@k for MZ, DZ and NT sibling identification using T1 data. MZ and DZ twins are divided into two groups based on median age of $29$. NT sibling pairs are divided based on median age difference of $3$ years.}
		\label{fig:Age_factor_impact_Twin_NT_Identify}
	}
\end{figure}

For the identification of MZ twins, we observe a mean recall near $100\%$ for all modalities, illustrating the high impact of genetic similarity on both structural and diffusion geometry in the brain. Comparing MZ, DZ and NT siblings, we see higher recall values for MZ twins compared to DZ twins or NT siblings, supporting the fact that MZ twins share more genetic information. We also note higher recall values for DZ twins compared to NT siblings, although both sibling types share the same genetic proximity. These differences are significant with $p < 0.01$ in a Wilcoxon signed-rank test. To rule out the impact of age difference, we further divided NT sibling pairs into two groups, based on the median age difference of $3$ years, and repeated the same experiment with T1 images. Figure \ref{fig:Age_factor_impact_Twin_NT_Identify} show no statistical difference between these two NT sub-groups. Similarly, there is no difference between MZ and DZ twins divided based on median age of $29$. 

Comparing individual modalities, we note that T1 leads to the highest recall values for identifying all sibling types. While T1 and T2 are highly correlated modalities, more feature matches were obtained for T1, suggesting this structural modality to be more relevant for the task of twin/sibling identification (see Figure \ref{fig:featMatchVis_NTpair_Structure_circles}). Moreover, the benefit of using a higher resolution is illustrated by the better performance of T1 compared to T1-FS, although these differences could also be explained by the lack of skull in T1-FS images. 

In general, diffusion measures (GFA, NQA0, NQA1) lead to smaller recall values than T1 or T2. This could be due to the fact that, for such images, the features are mainly located inside or near to white matter, as opposed to the whole brain for T1/T2 (see Figure \ref{fig:featMatchVis_NTpair_Structure_circles}). Comparing diffusion measures, GFA and NQA0 perform nearly the same, GFA having slightly higher recall values for MZ and NT sibling identification. This is possibly due to the fact that GFA contains information from the full diffusion profile, while NQA0 is based on the most prominent diffusion direction. 
NQA1, which contains information along the second most prominent diffusion direction, performs poorly compared to other diffusion measures.

Considering results obtained with multiple modalities, we note that the combination of T1 and T2 gives a higher recall than using these modalities individually. For instance, T1+T2 gives a recall@50 value of $0.706$ for NT identification, compared to $0.564$ and $0.549$ for T1 and T2, respectively. Similarly, combining T1 and all diffusion measures (i.e., T1+dMRI) improves the mean recall values significantly, for the identification of all sibling types (e.g., a recall@50 value of $0.722$ for NT identification). These results validate our hypothesis that different modalities provide complementary information.

\subsection{Scale-space visualization of features matches}
\label{subsubsec:Qualitative_compare_structure_data}

\begin{figure}[ht!]
	{\centering
	\begin{small}	
	   \begin{tabular}{ccc}
			\figViz{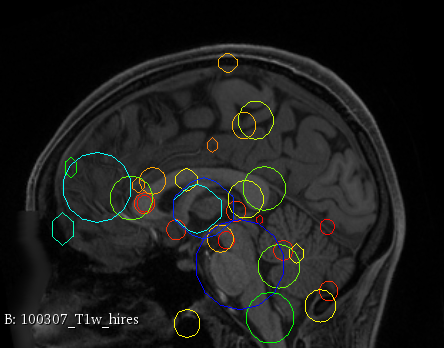} &
			\figViz{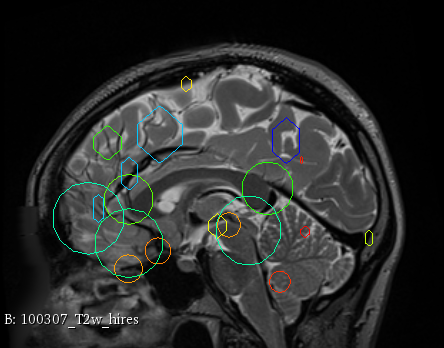} &
			\figViz{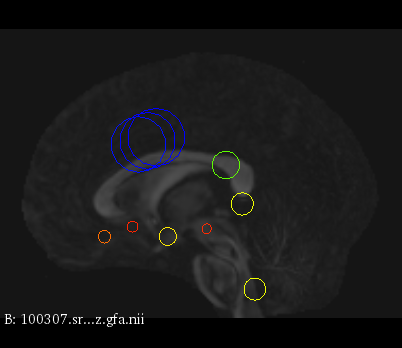} \\
			\figViz{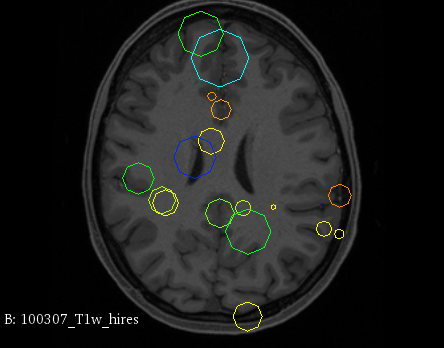} &
			\figViz{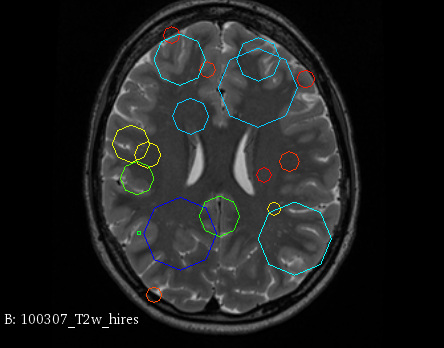} &
			\figViz{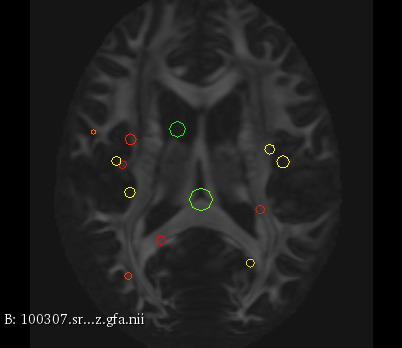} \\
			\figViz{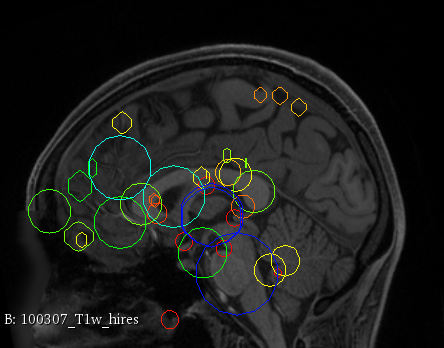} &
			\figViz{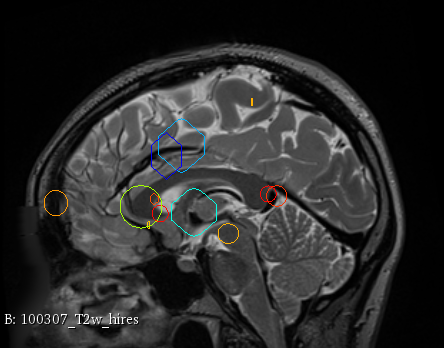} &
			\figViz{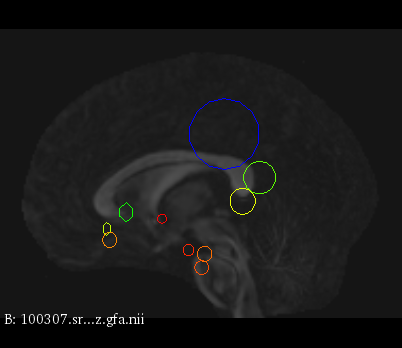} \\
			\shortstack{\figViz{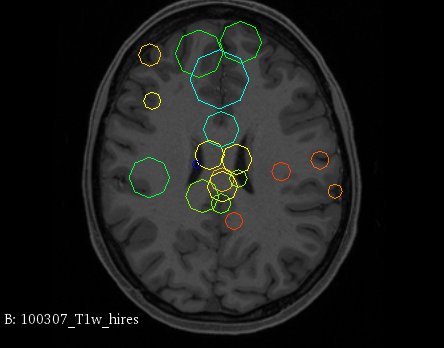} \\[1mm] T1} &
			\shortstack{\figViz{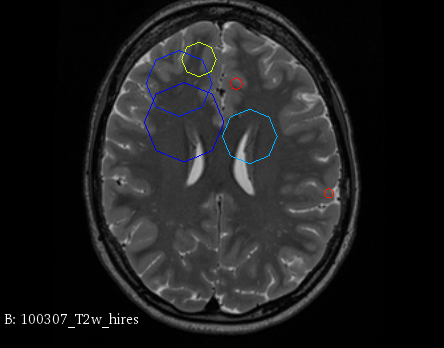} \\[1mm] T2} &
			\shortstack{\figViz{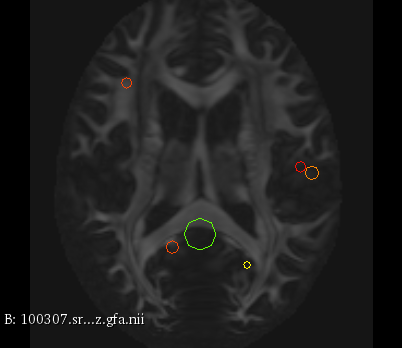} \\[1mm] GFA}
		\end{tabular}
		\end{small}
		\caption{Example of feature matches for a single pair of MZ twins (first and second row) and non-twin siblings (third and fourth row). Scale space is represented using circle radius (for the visible slice).}
		\label{fig:featMatchVis_NTpair_Structure_circles}
	}
\end{figure}

Figure \ref{fig:featMatchVis_NTpair_Structure_circles} provides a scale-space visualization of feature matches for a single pair of MZ twins and NT siblings, where scale information is represented using the circle radius. Note that circles represent the intersection of 3D spheres with the visible slice and, thus, non-intersecting features are hidden in this 2D visualization.  

It can be seen that different image modalities generally result in distinct, complementary feature correspondences throughout the brain, allowing a rich characterization of both anatomical and connectivity structure. In T1 and T2 images, features are mainly located in the frontal lobe, corpus callosum and brain stem. Smaller-scale features are also visible along various cortical regions, as well as in sub-cortical structures near the basal ganglia. Although highly correlated, T1 images show significantly more feature matches than T2 images. Moreover, images based on diffusion measures have less matches than in structural modalities. These matches are located mostly inside or near to white matter: larger-scale features in the corpus-callosum, and smaller-scale ones in the brain stem and along white matter bundles. While not shown in the figure, the set of matches found by combining two modalities (e.g., T1 + T2) generally corresponds to the union of those obtained with these individual modalities. 

Comparing different sibling types, we observe a greater number of matches between MZ twins than NT siblings. This observation, which is easier to visualize in T2 and GFA images, is consistent with other analyses on twin datasets. In terms of feature location and scale, no obvious pattern can be seen when comparing these two sibling types. However, a more detailed analysis would be required to validate this assertion.

\section{Conclusion}
\label{sec:Conclusion}

We presented a framework based on SIFT descriptors for the multi-modal analysis of genetically-related subjects. This framework represents each subject as an order-less sets (i.e., bag) of local invariant features, which can be extracted from different modalities. An efficient strategy, computing Jaccard similarity on a feature-to-feature nearest-neighbor graph, was proposed to evaluate the similarity between subjects having a different number of features. In a quantitative analysis, mean recall@$k$ was used to measure the relation between the proposed subject similarity and genetic proximity, and the contribution/complementarity of information from different MRI modalities. In agreement with other twin data analysis in the literature, results show mono-zygotic twins to be more similar than di-zygotic twins and non-twin siblings. Likewise, it was found that di-zygotic twins are more similar than non-twin siblings, and that this similarity is not entirely explained by age difference. Our analysis also showed structural modalities, in particular T1, to be more useful than modalities measuring diffusion. However, combining both structural and diffusion data leads to a better sibling identification.

This work could be extended by further investigating the differences, in terms of feature location and similarity, between di-zygotic twins and non-twin siblings. A deeper analysis of aging effects could also be performed, for instance, using longitudinal data. Such analysis would help understand the effect of neuroplasticity on individual brain characteristics.

\paragraph{Acknowledgements:} Data were provided in part by the Human Connectome Project, WU-Minn Consortium (Principal Investigators: David Van Essen and Kamil Ugurbil; 1U54MH091657) funded by the 16 NIH Institutes and Centers that support the NIH Blueprint for Neuroscience Research; and by the McDonnell Center for Systems Neuroscience at Washington University.


\bibliographystyle{splncs03}

\begin{footnotesize}
\bibliography{References_KD}

\begin{thebibliography}{10}
\providecommand{\url}[1]{\texttt{#1}}
\providecommand{\urlprefix}{URL }

\bibitem{finn2015functional}
Finn, E.S., Shen, X., Scheinost, D., Rosenberg, M.D., Huang, J., Chun, M.M.,
  Papademetris, X., Constable, R.T.: Functional connectome fingerprinting:
  identifying individuals using patterns of brain connectivity. Nature
  neuroscience  (2015)

\bibitem{geevarghese2015subcortical}
Geevarghese, R., Lumsden, D.E., Hulse, N., Samuel, M., Ashkan, K.: Subcortical
  structure volumes and correlation to clinical variables in parkinson's
  disease. Journal of Neuroimaging  25(2),  275--280 (2015)

\bibitem{glasser2013minimal}
Glasser, M.F., Sotiropoulos, S.N., Wilson, J.A., Coalson, T.S., Fischl, B.,
  Andersson, J.L., Xu, J., Jbabdi, S., Webster, M., Polimeni, J.R., et~al.: The
  minimal preprocessing pipelines for the human connectome project. Neuroimage
  80,  105--124 (2013)

\bibitem{goldman2013motor}
Goldman, S., O'Brien, L.M., Filipek, P.A., Rapin, I., Herbert, M.R.: Motor
  stereotypies and volumetric brain alterations in children with autistic
  disorder. Research in autism spectrum disorders  7(1),  82--92 (2013)

\bibitem{kochunov2015heritability}
Kochunov, P., Jahanshad, N., Marcus, D., Winkler, A., Sprooten, E., Nichols,
  T.E., Wright, S.N., Hong, L.E., Patel, B., Behrens, T., et~al.: Heritability
  of fractional anisotropy in human white matter: a comparison of human
  connectome project and enigma-dti data. Neuroimage  111,  300--311 (2015)

\bibitem{KUMAR2017242}
Kumar, K., Desrosiers, C., Siddiqi, K., Colliot, O., Toews, M.: Fiberprint: A
  subject fingerprint based on sparse code pooling for white matter fiber
  analysis. NeuroImage  158,  242 -- 259 (2017),
  \url{http://www.sciencedirect.com/science/article/pii/S1053811917305669}

\bibitem{mangin2004framework}
Mangin, J.F., Riviere, D., Cachia, A., Duchesnay, E., Cointepas, Y.,
  Papadopoulos-Orfanos, D., Scifo, P., Ochiai, T., Brunelle, F., Regis, J.: A
  framework to study the cortical folding patterns. Neuroimage  23,  S129--S138
  (2004)

\bibitem{miranda2014connectotyping}
Miranda-Dominguez, O., Mills, B.D., Carpenter, S.D., Grant, K.A., Kroenke,
  C.D., Nigg, J.T., Fair, D.A.: Connectotyping: model based fingerprinting of
  the functional connectome. PloS one  9(11),  e111048 (2014)

\bibitem{mueller2013individual}
Mueller, S., Wang, D., Fox, M.D., Yeo, B.T., Sepulcre, J., Sabuncu, M.R.,
  Shafee, R., Lu, J., Liu, H.: Individual variability in functional
  connectivity architecture of the human brain. Neuron  77(3),  586--595 (2013)

\bibitem{de2011atlasing}
de~Schotten, M.T., Bizzi, A., Dell'Acqua, F., Allin, M., Walshe, M., Murray,
  R., Williams, S.C., Murphy, D.G., Catani, M., et~al.: Atlasing location,
  asymmetry and inter-subject variability of white matter tracts in the human
  brain with {MR} diffusion tractography. Neuroimage  54(1),  49--59 (2011)

\bibitem{thompson2013genetics}
Thompson, P.M., Ge, T., Glahn, D.C., Jahanshad, N., Nichols, T.E.: Genetics of
  the connectome. Neuroimage  80,  475--488 (2013)

\bibitem{7493398}
Toews, M., Wells, W.M.: How are siblings similar? how similar are siblings?
  large-scale imaging genetics using local image features. In: 2016 IEEE 13th
  International Symposium on Biomedical Imaging (ISBI). pp. 847--850 (April
  2016)

\bibitem{toews2010feature}
Toews, M., Wells, W., Collins, D.L., Arbel, T.: Feature-based morphometry:
  Discovering group-related anatomical patterns. NeuroImage  49(3),  2318--2327
  (2010)

\bibitem{van2013wu}
Van~Essen, D.C., Smith, S.M., Barch, D.M., Behrens, T.E., Yacoub, E., Ugurbil,
  K., Consortium, W.M.H., et~al.: The wu-minn human connectome project: an
  overview. Neuroimage  80,  62--79 (2013)

\bibitem{wachinger2015brainprint}
Wachinger, C., Golland, P., Kremen, W., Fischl, B., Reuter, M., Initiative,
  A.D.N., et~al.: Brainprint: a discriminative characterization of brain
  morphology. NeuroImage  109,  232--248 (2015)

\bibitem{yeh2016connectometry}
Yeh, F.C., Badre, D., Verstynen, T.: Connectometry: A statistical approach
  harnessing the analytical potential of the local connectome. NeuroImage  125,
   162--171 (2016)

\bibitem{yeh2010generalized}
Yeh, F.C., Wedeen, V.J., Tseng, W.Y.I.: Generalized q-sampling imaging. IEEE
  transactions on medical imaging  29(9),  1626--1635 (2010)

\end{thebibliography}
\end{footnotesize}

\end{document}